\documentclass[runningheads]{llncs}
\usepackage[T1]{fontenc}
\usepackage{amsmath}
\usepackage{graphicx,verbatim}
\usepackage{amssymb}
\usepackage{multirow}
\usepackage{subcaption}
\usepackage{booktabs}
\usepackage{amssymb}
\usepackage{marvosym}
\begin{document}
\title{Bridging the Markerless Tracking Validation Gap: Intraoperative X-ray Workflow Integration}
\titlerunning{Integrating X-ray for Clinical Markerless Tracking}
%

\author{Connor Daly\inst{1}\textsuperscript{(\Letter)} \and
Salvatore Russo\inst{2} \and
Jinendra Ekanayake\inst{3} \and
Daniel S. Elson\inst{1}
\and
Ferdinando Rodriguez y Baena\inst{1}}
%
%
\institute{The Hamlyn Centre for Robotic Surgery, Imperial College London, London, UK, \email{cd1723@ic.ac.uk}\\ 
\and
 Department of Neurosurgery, Imperial College Healthcare NHS Trust, London, UK
 \and
Stanford University School of Medicine, Stanford University, California, USA
}

\authorrunning{C. Daly \textit{et al.}}

\maketitle              
\begin{abstract}
Markerless intraoperative tracking of spinal anatomy via RGB-D sensors is a promising alternative to invasive bone-anchored fiducials, yet no existing clinical dataset provides independent subsurface ground truth for validating these methods. We address this gap with a clinically viable pipeline that derives per-vertebra 6-DoF pose from routine intraoperative C-arm fluoroscopy, co-registered to the RGB-D camera frame through a multimodal fiducial array detectable in both imaging modalities. Critically, the pipeline requires no additional radiation, instrumentation, or deviation from standard surgical workflows, relying solely on the fluoroscopic images already acquired during routine level checks. A calibrated transform chain links the vertebral pose recovered via multi-view 2D–3D CT-to-fluoroscopy registration to the RGB-D coordinate frame, enabling direct comparison of any markerless registration estimate against a subsurface reference. Ex vivo validation against optically tracked ovine vertebrae confirms end-to-end accuracy of 2.06 ± 0.29 mm and 3.35 ± 0.93°, sufficient to serve as independent ground truth for methods with substantially larger errors. We demonstrate the pipeline's utility by evaluating two markerless registration strategies on clinical data spanning 11 vertebrae across 1,500 frames from three surgical sequences: the first such evaluation against an independent subsurface reference in a clinical setting. The proposed framework provides reusable infrastructure for benchmarking emerging markerless tracking methods on real surgical data.


\end{abstract}
\section{Introduction and Motivation}

Markerless intraoperative tracking of orthopedic anatomy via RGB-D sensors is a promising emerging direction for surgical guidance \cite{9646929,liebmann2024automatic}. Current surgical navigation systems depend on invasive bone-anchored fiducials to register (and subsequently track) preoperative imaging with the surgical scene \cite{Medtronic_StealthStation_S8,wilson2024image}, introducing added operative time and complication risk \cite{gao2024mazor}. Markerless approaches instead extract anatomically meaningful correspondences directly from intraoperative point clouds via learned segmentation, enabling registration and tracking without physical instrumentation \cite{9646929,liebmann2024automatic}. Daly \textit{et al.} \cite{daly2025towards} advanced this direction with the first clinical markerless tracking dataset for spine surgery. However, their ground truth is inherently approximate: alignment relies on surgeon-indicated surface landmarks refined against intraoperative topology, neither of which references the pose of the subsurface anatomy. For clinical adoption, markerless systems must be validated against a gold-standard measurement offering quantifiable translational and rotational error metrics \cite{balachandran2008accuracy}, a requirement no existing clinical dataset satisfies.
Cadaveric studies have obtained accurate vertebral pose through rigidly fixed optical tracking arrays \cite{liu2020automatic,liebmann2021spinedepth}. Liebmann \textit{et al.} \cite{liebmann2021spinedepth}, for example, fixed 3D-printed constructs to vertebral bodies via K-wires, achieving sub-2 mm accuracy but requiring anterior exposure incompatible with standard posterior approaches, additional tracking hardware within the operative field, and an intraoperative CT. These requirements are not permissible within a clinical observational study, leaving a methodological gap: there is currently no clinically viable means of obtaining precise vertebral pose ground truth for evaluating markerless registration in real surgical settings.

To address this, we present a pipeline that leverages routine 
intraoperative C-arm imaging to derive ground-truth vertebral 
pose within the RGB-D camera frame. Our contributions include:
\begin{enumerate}
\item A multimodal data capture and labelling pipeline that 
co-registers intraoperative RGB-D recordings with ground-truth 
vertebral poses derived from routine intraoperative fluoroscopy, 
requiring no additional radiation, no instrumentation within the surgical field, or deviation 
from standard surgical workflows.
\item The first quantitative evaluation of markerless spine 
registration methods on clinical data against an independent 
subsurface reference, reporting per-vertebra translational and 
rotational error across three surgical sequences.
\item A clinically grounded evaluation framework, validated \textit{ex vivo} to 2.06±0.29 mm and 3.35±0.93°, which can serve as reusable infrastructure for benchmarking markerless registration methods against an independent subsurface reference. Upon publication we will release CAD files for the intraoperative marker array and calibration phantom, along with the evaluation code and annotated data.
\end{enumerate}

This work reports initial results from a large-scale ongoing clinical data collection study; the pipeline and evaluation framework are designed to scale as further cases are captured, enabling progressively larger-scale benchmarking of markerless methods. 
\begin{figure}
\centering
\includegraphics[width=0.9\textwidth]{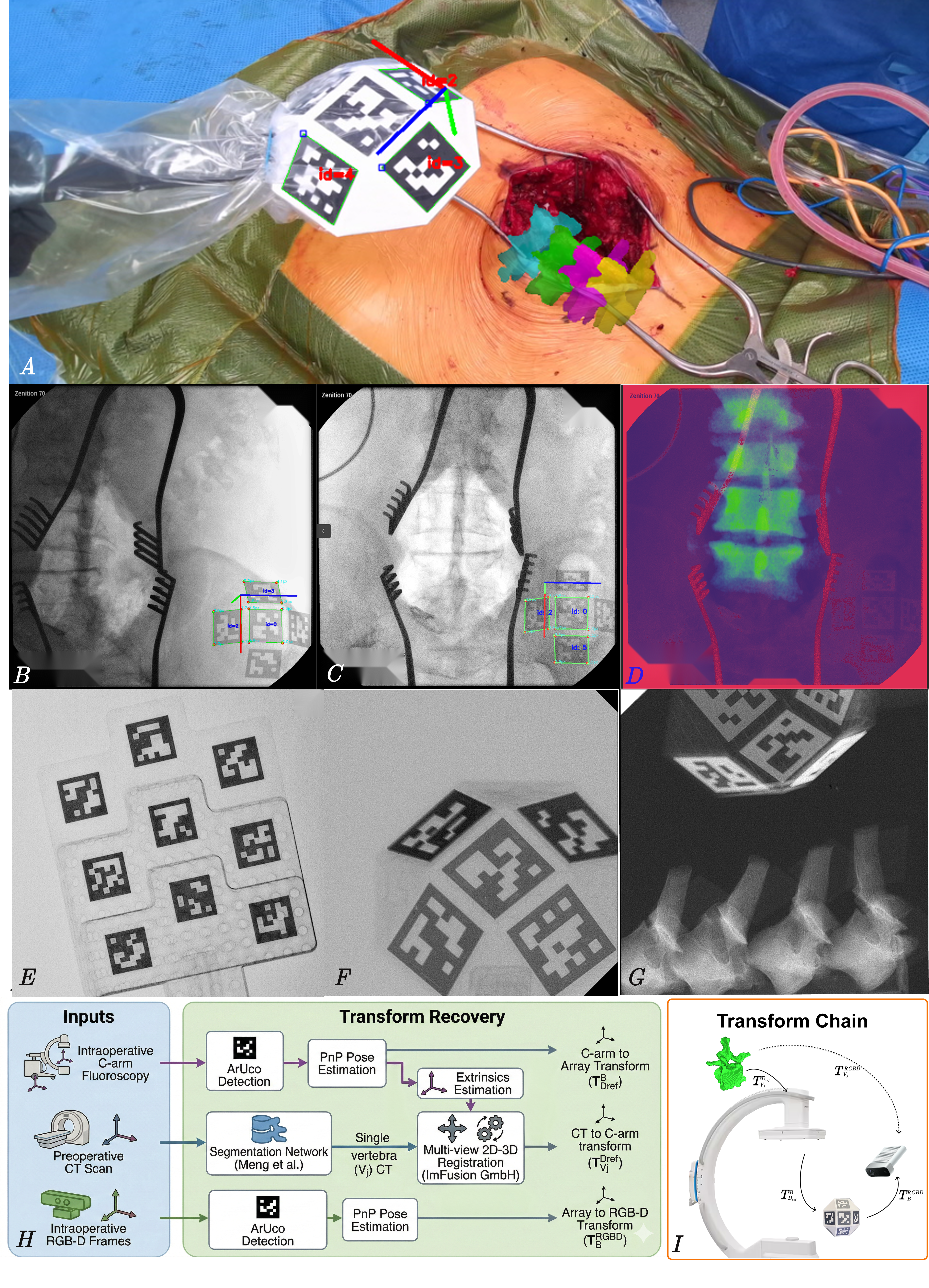}
\caption{\textbf{(A)} Intraoperative setup with the marker inside a sterile cover, secured taut with a rubber band by the scrub nurse to maintain detectability. The overlay shows automatic pose detection and projection of the preoperative CT into the surgical scene using the transform chain in \textbf{(I)}; \textbf{(B, C)} Automatic marker pose detection in C-arm images acquired during a fixation case, with projection errors (pixels) overlaid; \textbf{(D)} AP DRR images, showing CT registered to C-arm images in ImFusion; \textbf{(E)} Calibration phantom under fluoroscopy; \textbf{(F)} Marker array under fluoroscopy; \textbf{(G)} Marker array under fluoroscopy with ovine vertebrae; \textbf{(H)} Data processing overview, showing how transforms needed to project preoperative CT into the intraoperative RGB-D stream are acquired.
\textbf{(I)} Solid lines depict our method's full transform chain enabling projection of vertebra j’s CT into the Azure Kinect RGB-D camera frame (dashed line, $\mathbf{T}^{\mathbf{RGBD}}_{\mathbf{V}_{j}}$).}   \label{fig1}
\end{figure}
\section{Methodology}

\subsection{Data Capture}
In contrast to the use of coarse surface features to approximate the underlying pose of the spine by Daly \textit{et al.} \cite{daly2025towards}, we integrated routine intraoperative C-arm fluoroscopy, taken alongside the RGB-D recording, to locate the patient's preoperative anatomy within the intraoperative scene. C-arm fluoroscopy is broadly accessible across clinical settings \cite{chan2018essential,mackechnie2024establishing}, ensuring our pipeline generalizes across institutions without requiring intraoperative CT \cite{ji2025enhanced}. We used established multi-view 2D–3D registration algorithms, as implemented in the ImFusion software suite (ImFusion GmbH, Germany) \cite{imfusion2025suite}, requiring at least two fluoroscopic images from separate poses with known projection geometry.

We modeled each C-arm source as a pinhole camera \cite{navab2010camera,gorges2005model,liu20152d} and calibrated intrinsics using a custom multi-plane phantom \cite{cho2005accurate,daly2008geometric} with nine radio-opaque ArUco markers \cite{garrido2014automatic} at known depths, estimating parameters via bundle adjustment \cite{opencv_library,navab2010camera}. Calibration images were acquired prior to surgery from AP and oblique angles within ±30°, reflecting the range of intraoperative C-arm positioning.

\subsection{Intraoperative Capture}
\label{intraoperative capture}
This study received approval from the relevant national research 
ethics authority (reference withheld for anonymization).
The cases we captured within our surgical study were taken from the clinical caseload of a tertiary neurosurgery department. The captured procedures are all lumbar-thoracic fixation cases. For recording RGB-D sequences, we used an Azure Kinect camera (Microsoft 
Corporation, USA), positioned directly over the incision using a boom arm fixed to a tripod. 

To co-register fluoroscopic and RGB-D coordinate frames without external tracking hardware, we designed a multimodal fiducial array, automatically 
detectable in both imaging modalities. Five radiopaque ArUco markers, each 30 $\times$ 30 mm, are inset on a radio-translucent PETG body. The insets were 3D-printed using PLA XRS filament (Nanovia Tech, France). The non-coplanar array layout ensured no marker overlap occurs at either AP or oblique viewing angles. The array was held on a boom arm adjacent to the exposition site, beneath the C-arm detector and within the RGB-D camera field of view, and covered with an ultrasound probe cover to maintain the sterile field.

An intraoperative C-arm is routinely used to provide confirmation of the vertebra level being operated on, typically from an AP view with adjacent oblique views. As the marker array was adjacent to the exposition, it is visible in the X-ray field of view during these level check shots, allowing both the determination of relative C-arm pose and the co-registration of X-ray and RGB-D coordinate frames to be performed with the same array. For 
each clinical case, we also collected the accompanying preoperative CT scan. We segmented the relevant bone tissue from CT using the network developed by Meng \textit{et al.} \cite{meng2023vertebrae}.

\subsection{CT Imaging to RGB-D Registration}
\label{methodo}
For each C-arm shot, using the acquired intrinsic parameters from our multi-view calibration, the radiopaque ArUco markers are automatically detected within the image, and we then recovered the board-to-detector transform via PnP pose estimation \cite{fischler1981random,jain2005ftrac}. Designating the first acquired image as the reference frame, we computed relative transforms between shots through the common marker array coordinate system, and combined these with the previously calibrated intrinsic parameters to yield projection matrices suitable for multi-view 2D–3D registration. 
Multi-view 2D-3D X-ray to CT registration is a well-established technique capable of achieving sub-millimeter accuracy \cite{markelj2012review}. We then performed intensity-based 2D–3D registration, optimizing over six rigid-body parameters by maximizing mutual information between digitally reconstructed radiographs (DRRs) and the acquired fluoroscopic images, using the ImFusion software suite.

This process was repeated for each segmented vertebra $\mathbf{V}_{j}$, yielding the transform from $\mathbf{V}_{j}$ to the detector coordinate frame, $\mathbf{D}_{ref}$, at the reference position. 
The transforms $\mathbf{T}^{\mathbf{D}_{ref}}_{\mathcal{B}}$ and $\mathbf{T}^{\mathbf{RGBD}}_{\mathcal{B}}$, locating the marker array in the reference detector frame and the RGB-D camera frame respectively, were both recovered via ArUco detection and PnP pose estimation.

Thus, we obtained the complete transform chain from the coordinate system of the vertebra $\mathbf{V}_{j}$ to the camera $\mathbf{C}$:
$$\mathbf{T}^{\mathbf{RGBD}}_{\mathbf{V}_{j}} = \mathbf{T}^{\mathbf{RGBD}}_{\mathcal{B}}  \cdot {(\mathbf{T}^{\mathbf{D}_{ref}}_{\mathcal{B}})}^{-1} \cdot \mathbf{T}^{\mathbf{D}_{ref}}_{\mathbf{V}_{j}}$$

\section{Evaluation and Results}
\subsection{\textit{Ex Vivo} Validation}
\label{method_resukt}
To demonstrate the accuracy and repeatability of our data collection and labeling pipeline, we conducted several validation experiments. We first established the level of error introduced by each individual component of the transform chain. To do this, we first rigidly fixed a passive infrared tracking array to the multimodal marker array. We then directly compared how well the marker array itself could be tracked in both the C-arm imaging, as well as the Azure Kinect RGB frames, with respect to a highly accurate optical tracker (Polaris Vicra, Northern Digital Inc, Canada; 0.25\,mm RMS volumetric accuracy, 0.50\,mm 95\% CI) \cite{ndi_polaris_vicra}. For each imaging system, the marker array was moved between $4$ distinct points in front of the sensor. We present the results in Tab. \ref{tab:component_validation}.

\begin{table}
\centering
\caption{Independent validation of each coordinate recovery stage 
against optical tracker ground truth. RGB-D tracking is evaluated 
with the sterile probe cover in place. Transitions 1--3 represent 
lateral displacement within the image plane; transition 4 represents 
displacement along the sensor optical axis.}
\begin{tabular}{l|cc|cc|cc}
\hline
& \multicolumn{2}{c|}{$1 \rightarrow 2$} & \multicolumn{2}{c|}{$1 \rightarrow 3$} & \multicolumn{2}{c}{$1 \rightarrow 4$} \\
\textbf{Modality} & \textbf{Trans.} & \textbf{Rot.} & \textbf{Trans.} & \textbf{Rot.} & \textbf{Trans.} & \textbf{Rot.} \\
& (mm) & ($^\circ$) & (mm) & ($^\circ$) & (mm) & ($^\circ$) \\
\hline
RGB-D tracking         & 0.40 & 0.31 & 0.70 & 0.61 & 0.10 & 1.85 \\
X-ray tracking (AP)    & 0.58 & 0.03 & 1.19 & 0.07 & 0.02 & 0.35 \\
X-ray tracking (oblique) & 0.46 & 0.18 & 0.96 & 0.50 & 0.66 & 0.38 \\
\hline
\end{tabular}
\label{tab:component_validation}
\end{table}

Subsequently, to validate the end-to-end accuracy of the full transform chain, we compared the difference in pose of four excised ovine vertebrae determined by our pipeline, with that found by directly tracking ArUco markers, rigidly fixed to the vertebrae. We mounted two plastic ArUco markers to each of the vertebrae with plastic screws, minimizing the markers' visibility within the C-arm imaging. We then performed CT scans of each vertebra. Within the coordinate space of the generated scans, we then localized the ArUco positions, meaning then that each vertebra could be tracked directly via the RGB-D camera, by-passing the transform chain shown in Fig. \ref{fig1} (I) entirely.

To compute registration differences, we selected two points (the tip of the spinous process, and a point on the superior laminar margin) for establishing a Total Registration Error (TRE), with the optically tracked pose representing the target. 
 The vertebrae and radio-opaque marker array were positioned between the source and detector of a TAU 2020 C-arm (Ziehm-Orthoscan Inc, USA), as described in section \ref{intraoperative capture}. With the radio-opaque marker array visible from the RGB-D camera, we captured 2 C-arm shots with an angular disparity of $30^{\circ}$. This is comparable to the mean angular disparity of $25.6^{\circ}$ seen in clinical cases we captured. Following the process described in \ref{methodo}, we projected the vertebrae from the CT coordinate space to the RGB-D camera coordinate space.
We present the TRE for the point landmarks as well as the rotational difference in Tab. \ref{tab:extab1}.
\begin{table}[ht]
\centering
\caption{\textit{Ex vivo} registration accuracy across ovine thoracic vertebrae. Combined TRE represents the mean of both anatomical landmark measurements.}
\begin{tabular}{c|cc}
\hline
\textbf{Vertebra} & \textit{Combined TRE (mm)} & \textit{Rotational Error ($^\circ$)} \\
\hline
T1 & 1.96 & 4.06 \\
T2 & 2.32 & 3.40 \\
T3 & 1.62 & 3.94 \\
T4 & 2.32 & 1.98 \\
\hline
\textbf{Agg.} & \textbf{2.06 $\pm$ 0.29} & \textbf{3.35 $\pm$ 0.93} \\
\hline
\end{tabular}
\label{tab:extab1}
\end{table}
\begin{figure}
\centering
\includegraphics[width=0.6\textwidth]{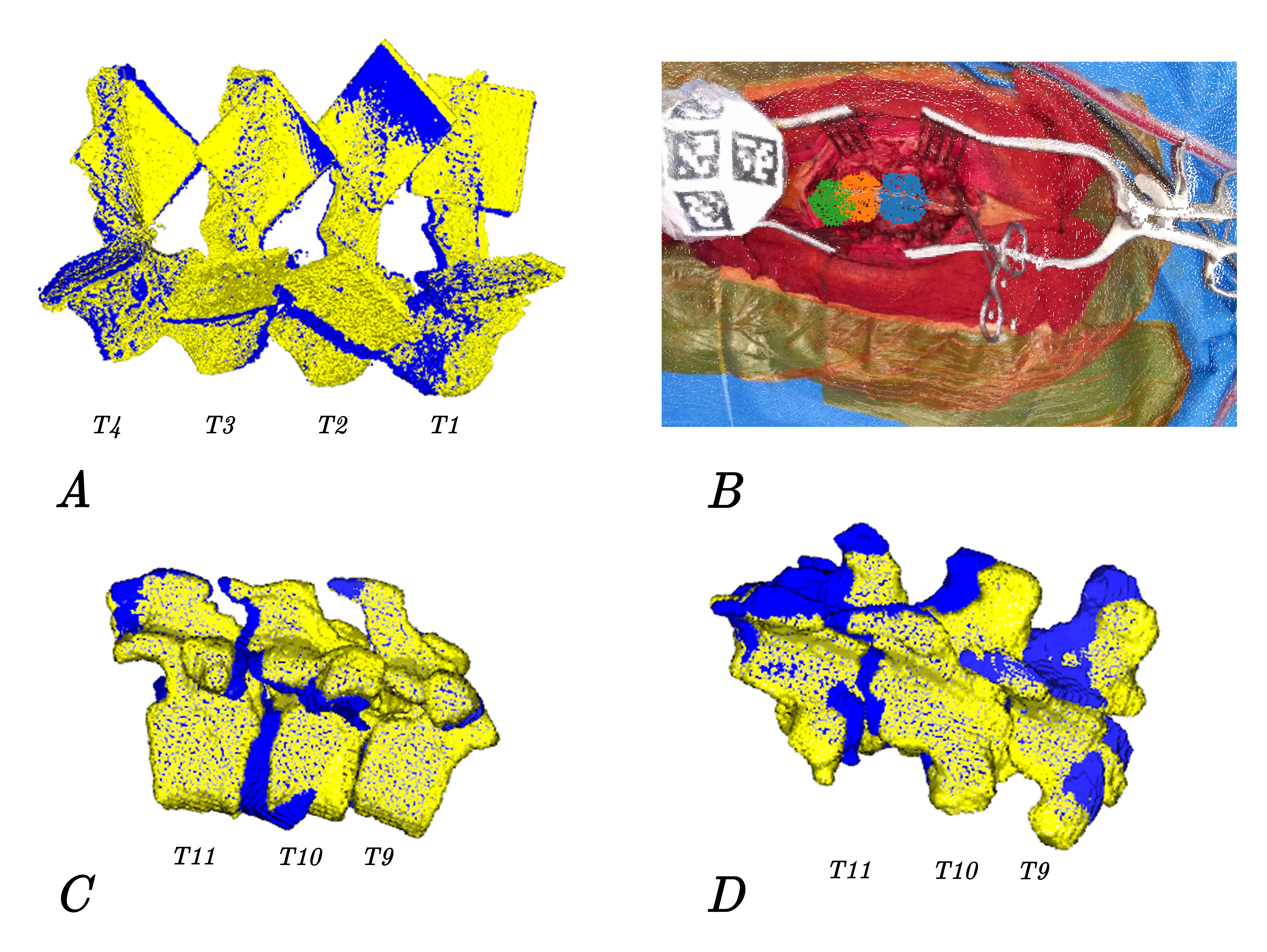}
\caption{\textbf{(A)} The pose of ovine CT meshes directly tracked via the Azure Kinect in blue, compared to those tracked via fluoroscopy and projected into the Kinect camera space in yellow; \textbf{(B)} Per vertebra patches predicted by the network in Sec. \ref{clinical}; \textbf{(C, D)} Intraoperative pose estimates via X-ray in yellow, and through per vertebra ICP in blue.}\label{figsheep}
\end{figure}
The independent validation of the RGB-D and X-ray pose tracking against an optical tracker reference confirms sub-millimeter translational agreement and sub-degree rotational agreement. The end-to-end pipeline, validated \textit{ex vivo} against fiducial-tracked vertebral pose, yields a combined TRE of $2.06 \pm 0.29$ mm and rotational error of $3.35 \pm 0.93^{\circ}$, approaching the $\leq 2$ mm and $\leq 5^{\circ}$ thresholds commonly cited for spinal navigation systems \cite{miller2016reliability}. Critically, this level of precision is sufficient to serve as independent ground truth for evaluating markerless tracking methods, whose registration errors are substantially larger, particularly in rotation.

\subsection{Clinical Data Registration}
\label{clinical}

With the pipeline validated in the lab, we demonstrated its utility by evaluating two markerless registration strategies on clinical data from three surgical sequences, encompassing 11 vertebrae across 1,500 frames. To our knowledge, this represents the first evaluation of markerless spine registration against an independent subsurface gold standard reference in a clinical setting.

The first strategy follows \cite{daly2025towards}, deforming the preoperative spine mesh via a kinematic chain model before rigidly registering it to segmented intraoperative points using ICP. This approach relies on semi-automated surface-based optimization of the kinematic chain parameters, with manual refinement, to align the deformed mesh with the intraoperative point cloud: a workflow difficult to scale to real-time intraoperative use. We therefore additionally trialed a fully automated approach inspired by \cite{liebmann2024automatic}, in which CT meshes of individual vertebrae are independently registered to per-vertebra segmentation patches. These patches were obtained via a two-stage pipeline: a PointNet++ \cite{qi2017pointnet++} heatmap regression network, trained on manually annotated spinous process tip positions, first localizes tips within the point cloud (recall 0.95, F1 0.89, localization error $3.98 \pm 1.99$\,mm), after which a PointNet segmentation network isolates points within 40\,mm of each detection (IoU 0.71, F1 0.83). Per-vertebra segmentation labels were generated by computing mesh-to-point-cloud proximity for individual vertebral CT meshes from \cite{daly2025towards}, following the same approach used therein for global spine masks. The results of both registration approaches, evaluated against our fluoroscopic pose estimates with a correspondence distance of 3\,mm for both methods, are shown in Tab.~\ref{tab:registration_comparison}.

\begin{table}[ht]
\centering
\caption{Comparison of \cite{daly2025towards}'s rigid registration method across three sequences of clinical data, versus individual ICP. Translation (Trans.) errors are reported in mm and rotation (Rot.) errors in degrees. Each sequence is 500 frames. "Exposure" represents the exposed vertebra count.}
\label{tab:registration_comparison}
\begin{tabular}{c|cc|cc}
\hline
\textbf{Exposure} 
& \multicolumn{2}{c|}{\textbf{Per Vertebra ICP}} 
& \multicolumn{2}{c}{\textbf{Rigid ICP}} \\

& Trans. (mm) & Rot. ($\circ$) 
& Trans. (mm) & Rot. ($\circ$) \\
\hline
5 & $5.15 \pm 3.65$ & $8.21 \pm 3.74$ & $5.70 \pm 1.59$ & $4.61 \pm 1.55$ \\
3 & $3.98 \pm 1.81$ & $9.10 \pm 3.68$ & $4.66 \pm 0.82$ & $3.53 \pm 0.78$ \\
3 & $4.02 \pm 3.10$ & $7.86 \pm 3.63$ & $3.38 \pm 0.58$ & $5.36 \pm 0.30$ \\
\hline
\textbf{Overall} 
& \textbf{4.52 $\pm$ 3.00} 
& \textbf{8.36 $\pm$ 3.69} 
& \textbf{4.78 $\pm$ 1.10} 
& \textbf{4.52 $\pm$ 1.00} \\
\hline
\end{tabular}
\end{table}

The rigid method of \cite{daly2025towards} outperforms per-vertebra ICP on rotation ($4.52^{\circ} \pm 1.00^{\circ}$ vs.\ $8.36^{\circ} \pm 3.69^{\circ}$) and exhibits substantially lower variance across all metrics, while mean translation errors are comparable with overlapping distributions. The per-vertebra approach, though fully automated, lacks the geometric constraints afforded by registering the full spine as a single structure, limiting its ability to resolve orientation. While both methods achieve mean translational errors below 5\,mm, neither yet meets the $\leq 2$ mm and $\leq 5^{\circ}$ thresholds cited for spinal navigation \cite{miller2016reliability}, underscoring both the value of independent pose evaluation and the headroom remaining for emerging methods.
\section{Discussion}

The clinical registration comparison in Tab. \ref{tab:registration_comparison} demonstrates the core contribution of this work: for the first time, competing markerless registration strategies can be evaluated on real surgical data against an independent subsurface reference. Prior clinical evaluation was limited to surrogate surface metrics \cite{daly2025towards}; our pipeline provides per-vertebra 6-DoF pose estimates from routine intraoperative fluoroscopy without additional hardware or workflow disruption.

 Registration errors in Tab. \ref{tab:registration_comparison} are measured against our fluoroscopic estimates, which carry an intrinsic uncertainty of $2.06 \pm 0.29$\,mm and $3.35 \pm 0.93^{\circ}$ (Tab.~\ref{tab:extab1}); these values thus represent upper bounds on true registration error. Alas, the rotational difference between the two methods ($ 3.84^{\circ}$) exceeds our reference uncertainty, confirming that the rigid method of \cite{daly2025towards} is meaningfully more accurate in rotation. Translational errors are not statistically different. Overall, the rigid approach is clearly superior, but its reliance on manual deformation tuning remains a barrier to clinical deployment, and thus an avenue for future work.

Key limitations remain. The study reports three clinical sequences and ovine \textit{ex vivo} validation; further labeled cases, drawn from a larger ongoing clinical study, will broaden the evaluation. Accuracy is also constrained by the 2D–3D registration stage, which depends on image quality, angular separation, and the number of fluoroscopic views.

Clinically, classical surface registration is intrinsically limited: the posterior vertebral surface exposed during surgery is geometrically ambiguous, leaving ICP unable to robustly resolve pose, particularly rotation. This is unlikely to improve with initialization alone. With sufficient fluoroscopic ground truth, a learned mapping from intraoperative point clouds to vertebral pose could instead leverage appearance, anatomical context, and priors beyond ICP’s scope.

\begin{credits}

\end{credits}

%
%
%
%

\bibliographystyle{splncs04} 
\bibliography{bib}

@article{navab2010camera,
  title={Camera augmented mobile C-arm (CAMC): calibration, accuracy study, and clinical applications},
  author={Navab, Nassir and Heining, Sandro-Michael and Traub, Joerg},
  journal={IEEE Transactions on Medical Imaging},
  volume={29},
  number={7},
  pages={1412--1423},
  year={2010}
}

@article{garrido2014automatic,
  title={Automatic generation and detection of highly reliable fiducial markers under occlusion},
  author={Garrido-Jurado, Sergio and Mu{\~n}oz-Salinas, Rafael and Madrid-Cuevas, Francisco Jos{\'e} and Mar{\'\i}n-Jim{\'e}nez, Manuel Jes{\'u}s},
  journal={Pattern Recognition},
  volume={47},
  number={6},
  pages={2280--2292},
  year={2014}
}

@article{daly2008geometric,
  title={Geometric calibration of a mobile C-arm for intraoperative cone-beam CT},
  author={Daly, Michael J and Siewerdsen, Jeffrey H and Cho, Young B and Jaffray, David A and Irish, Jonathan C},
  journal={Medical physics},
  volume={35},
  number={5},
  pages={2124--2136},
  year={2008}
}

@inproceedings{daly2025towards,
  title={Towards Markerless Intraoperative Tracking of Deformable Spine Tissue},
  author={Daly, Connor and Marconi, Elettra and Riva, Marco and Ekanayake, Jinendra and Elson, Daniel S and Rodriguez y Baena, Ferdinando},
  booktitle={International Conference on Medical Image Computing and Computer-Assisted Intervention},
  pages={627--637},
  year={2025},
  organization={Springer}
}

@article{meng2023vertebrae,
  title={Vertebrae localization, segmentation and identification using a graph optimization and an anatomic consistency cycle},
  author={Meng, Di and Boyer, Edmond and Pujades, Sergi},
  journal={Computerized Medical Imaging and Graphics},
  volume={107},
  pages={102235},
  year={2023},
  publisher={Elsevier}
}

@article{jain2005ftrac,
  title={FTRAC—A robust fluoroscope tracking fiducial},
  author={Jain, Ameet Kumar and Mustafa, Tabish and Zhou, Yu and Burdette, Clif and Chirikjian, Gregory S and Fichtinger, Gabor},
  journal={Medical physics},
  volume={32},
  number={10},
  pages={3185--3198},
  year={2005},
  publisher={Wiley Online Library}
}

@article{fischler1981random,
  title={Random sample consensus: a paradigm for model fitting with applications to image analysis and automated cartography},
  author={Fischler, Martin A and Bolles, Robert C},
  journal={Communications of the ACM},
  volume={24},
  number={6},
  pages={381--395},
  year={1981},
  publisher={ACM New York, NY, USA}
}

@article{markelj2012review,
  title={A review of 3D/2D registration methods for image-guided interventions},
  author={Markelj, Primoz and Toma{\v{z}}evi{\v{c}}, Dejan and Likar, Bostjan and Pernu{\v{s}}, Franjo},
  journal={Medical image analysis},
  volume={16},
  number={3},
  pages={642--661},
  year={2012},
  publisher={Elsevier}
}

@article{miller2016reliability,
  title={Reliability of the planned pedicle screw trajectory versus the actual pedicle screw trajectory using intra-operative 3D CT and image guidance},
  author={Miller, Catherine A and Ledonio, Charles G and Hunt, Matthew A and Siddiq, Farhan and Polly, David W},
  journal={International Journal of Spine Surgery},
  volume={10},
  year={2016},
  publisher={International Journal of Spine Surgery}
}

@article{chan2018essential,
  title={Essential fracture and orthopaedic equipment lists in low resource settings: consensus derived by survey of experts in Africa},
  author={Chan, Yuen and Banza, Leonard and Martin Jr, Claude and Harrison, William J},
  journal={BMJ open},
  volume={8},
  number={9},
  pages={e023473},
  year={2018},
  publisher={British Medical Journal Publishing Group}
}

@article{mackechnie2024establishing,
  title={Establishing consensus on essential resources for musculoskeletal trauma care worldwide: a modified Delphi study},
  author={MacKechnie, Madeline C and Shearer, David W and Verhofstad, Michael HJ and Martin, Claude and Graham, Simon M and Pesantez, Rodrigo and Schuetz, Michael and H{\"u}ttl, Tobias and Kojima, Kodi and Bernstein, Brian P and others},
  journal={JBJS},
  volume={106},
  number={1},
  pages={47--55},
  year={2024},
  publisher={LWW}
}

@article{ji2025enhanced,
  title={Enhanced Safety and Efficacy of O-Arm Navigation Over C-Arm Guidance in Percutaneous Kyphoplasty for Patients With Osteoporotic Vertebral Compression Fractures and Spinal Deformity: A Comparative Study},
  author={Ji, Zhong-Wei and Shen, Min-Jie and Sun, Jia-Jia and Wang, Jia-Le and Deng, Yong-Kang and Zhang, Yao and Wu, Xie-Xing and Chen, Kang-Wu and Mao, Hai-Qing},
  journal={Orthopaedic Surgery},
  volume={17},
  number={1},
  pages={125--132},
  year={2025},
  publisher={Wiley Online Library}
}

@inproceedings{gorges2005model,
  title={Model of a vascular C-arm for 3D augmented fluoroscopy in interventional radiology},
  author={Gorges, S{\'e}bastien and Kerrien, Erwan and Berger, Marie-Odile and Trousset, Yves and Pescatore, J{\'e}r{\'e}mie and Anxionnat, Ren{\'e} and Picard, Luc},
  booktitle={International Conference on Medical Image Computing and Computer-Assisted Intervention},
  pages={214--222},
  year={2005},
  organization={Springer}
}

@article{liu20152d,
  title={2D--3D radiograph to cone-beam computed tomography (CBCT) registration for C-arm image-guided robotic surgery},
  author={Liu, Wen Pei and Otake, Yoshito and Azizian, Mahdi and Wagner, Oliver J and Sorger, Jonathan M and Armand, Mehran and Taylor, Russell H},
  journal={International journal of computer assisted radiology and surgery},
  volume={10},
  number={8},
  pages={1239--1252},
  year={2015},
  publisher={Springer}
}

@article{cho2005accurate,
  title={Accurate technique for complete geometric calibration of cone-beam computed tomography systems},
  author={Cho, Youngbin and Moseley, Douglas J and Siewerdsen, Jeffrey H and Jaffray, David A},
  journal={Medical physics},
  volume={32},
  number={4},
  pages={968--983},
  year={2005},
  publisher={Wiley Online Library}
}

@misc{opencv_library,
  author       = {Bradski, Gary and Kaehler, Adrian},
  title        = {OpenCV: Open Source Computer Vision Library},
  year         = {2000},
  howpublished = {\url{https://opencv.org/}},
  note         = {Accessed: 2026-02-08}
}

@ARTICLE{9646929,
  author={Hu, Xue and Nguyen, Anh and Baena, Ferdinando Rodriguez y},
  journal={Transactions on Instrumentation and Measurement}, 
  title={Occlusion-Robust Visual Markerless Bone Tracking for Computer-Assisted Orthopedic Surgery}, 
  year={2022},
  volume={71},
  number={},
  pages={1-11},
  doi={10.1109/TIM.2021.3134764}}

@article{liebmann2024automatic,
  title={Automatic registration with continuous pose updates for marker-less surgical navigation in spine surgery},
  author={Liebmann, Florentin and von Atzigen, Marco and St{\"u}tz, Dominik and Wolf, Julian and Zingg, Lukas and Suter, Daniel and Cavalcanti, Nicola A and Leoty, Laura and Esfandiari, Hooman and Snedeker, Jess G and others},
  journal={Medical Image Analysis},
  volume={91},
  pages={103027},
  year={2024},
  publisher={Elsevier}
}

@misc{Medtronic_StealthStation_S8,
key = {s},
  title = {StealthStation S8 Surgical Navigation System},
  howpublished = "\url{https://europe.medtronic.com/xd-en/healthcare-professionals/products/neurological/surgical-navigation-systems/stealthstation/stealthstation-s8.html}",
  note = {Accessed: 2026-02-26}
}

@article{wilson2024image,
  title         = {
    Image-guided navigation in spine surgery: from historical developments to
    future perspectives
  },
  author        = {
    Wilson Jr, John Preston and Fontenot, Lane and Stewart, Caleb and Kumbhare,
    Deepak and Guthikonda, Bharat and Hoang, Stanley
  },
  year          = {2024},
  journal       = {Journal of Clinical Medicine},
  publisher     = {MDPI},
  volume        = {13},
  number        = {7},
  pages         = {2036}
}

@article{gao2024mazor,
  title         = {
    Mazor X robot-assisted upper and lower cervical pedicle screw fixation: a
    case report and literature review
  },
  author        = {
    Gao, Zilin and Zhang, Xiaobo and Xu, Zhengwei and Jiang, Chao and Hu, Wei and
    Zhang, Haiping and Hao, Dingjun
  },
  year          = {2024},
  journal       = {BMC Geriatrics},
  publisher     = {Springer},
  volume        = {24},
  number        = {1},
  pages         = {916}
}

@article{balachandran2008accuracy,
  title={Accuracy of image-guided surgical systems at the lateral skull base as clinically assessed using bone-anchored hearing aid posts as surgical targets},
  author={Balachandran, Ramya and Fitzpatrick, J Michael and Labadie, Robert F},
  journal={Otology \& Neurotology},
  volume={29},
  number={8},
  pages={1050--1055},
  year={2008},
  publisher={LWW}
}

@article{liu2020automatic,
  title         = {
    Automatic markerless registration and tracking of the bone for
    computer-assisted orthopaedic surgery
  },
  author        = {Liu, He and Baena, Ferdinando Rodriguez Y},
  year          = {2020},
  journal       = {IEEE Access},
  publisher     = {IEEE},
  volume        = {8},
  pages         = {42010--42020}
}

@article{liebmann2021spinedepth,
  title={Spinedepth: a multi-modal data collection approach for automatic labelling and intraoperative spinal shape reconstruction based on rgb-d data},
  author={Liebmann, Florentin and St{\"u}tz, Dominik and Suter, Daniel and Jecklin, Sascha and Snedeker, Jess G and Farshad, Mazda and F{\"u}rnstahl, Philipp and Esfandiari, Hooman},
  journal={Journal of Imaging},
  volume={7},
  number={9},
  pages={164},
  year={2021},
  publisher={MDPI}
}

@article{qi2017pointnet++,
  title         = {
    Pointnet++: Deep hierarchical feature learning on point sets in a metric
    space
  },
  author        = {Qi, Charles Ruizhongtai and Yi, Li and Su, Hao and Guibas, Leonidas J},
  year          = {2017},
  journal       = {Advances in neural information processing systems},
  volume        = {30}
}

@misc{imfusion2025suite,
    author = {ImFusion},
  title        = {{ImFusion Suite}: Visual Computing for Medical Imaging},
  howpublished = {\url{https://imfusion.com/products-overview/imfusion-suite/}},
  year         = {2025},
  note         = {Accessed: 2025-11-14},
  organization = {ImFusion}
}

@manual{ndi_polaris_vicra,
  title        = {{Polaris Vicra Technical Specifications}},
  author       = {{Northern Digital Inc.}},
  year         = {2018},
  note         = {P/N 10001248 (Rev 001)},
  url          = {https://www.ndigital.com/products/polaris-vicra/}
}
\end{document}